\documentclass{article}
\usepackage{arxiv}

\usepackage{latexsym}
\usepackage{amssymb}
\usepackage{amsmath}
\usepackage{booktabs}
\usepackage{enumitem}
\usepackage{graphicx}
\usepackage{color}
\usepackage{times,soul,url,algorithm,algorithmic,xcolor,comment,amsfonts}
\usepackage[switch]{lineno}
\usepackage{tikz}
\tikzstyle{densely dotted}= [dash pattern=on \pgflinewidth off 1pt]
\usetikzlibrary{decorations.text}
\usetikzlibrary{positioning}
\usetikzlibrary{decorations.pathmorphing}
\usetikzlibrary{arrows,automata,decorations}
\usepackage{subfigure}

\newtheorem{example}{Example}
\newtheorem{definition}{Definition}

\begin{document}

\title{Frank's triangular norms \\in Piaget's logical proportions}

%
\author{{Henri Prade} \\
Institut de Recherche en Informatique de Toulouse (IRIT) - CNRS,\\ 118, route de Narbonne, 31062  Toulouse Cedex 9, France \\
\texttt{henri.prade@irit.fr}
\And{Gilles Richard}\\
Institut de Recherche en Informatique de Toulouse (IRIT),\\ 118, route de Narbonne, 31062  Toulouse Cedex 9, France \\
\texttt{gilles.richard@irit.fr}
%
%
}


\maketitle
\begin{abstract}Starting from the Boolean notion of logical proportion in Piaget's sense, which turns out to be equivalent to analogical proportion, this note proposes a definition of analogical proportion between numerical values based on triangular norms (and dual co-norms). Frank's family of triangular norms is particularly interesting from this perspective. The article concludes with a comparative discussion with another very recent proposal for defining analogical proportions between numerical values based on the family of generalized means. 

\end{abstract}
\section{Introduction}
In the appendix to a 1952 French book \cite{Piaget1952}, in which he continued his systematic study of the transformations of logical statements, psychologist Jean Piaget introduced a quaternary operation he called \emph{logical proportion}. He considered the notion important enough to devote a few pages to it in his introductory book \cite{Piaget1953} (pp. 35--37) the following year.  It turns out that this so-called \emph{logical proportion} is one of the possible (equivalent) expressions of a Boolean analogical proportion \cite{PraRicLU2013}.  However, Piaget apparently never made the connection between his logical proportions and analogy.

This is the starting point of this short paper for proposing a new extension of analogical proportions to numerical values, based on triangular norms \cite{KMP00}.

The paper is organized as follows. \begin{itemize}
    \item 
Section \ref{Piaget}
recalls the link between Piaget's logical proportions and Boolean analogical proportions. 

   \item Section \ref{triang}  presents a  new way of extending analogical proportions to numerical data  using a pair of dual triangular norm and co-norm in place of conjunction and disjunction connectives  in Piaget's logical proportions.  This preserves the expected properties of analogical proportions. Moreover  the Frank's family of triangular norms and co-norms \cite{Frank79} seems especially appropriate, since it preserves an additional  desirable  property. 

   \item Section \ref{related}   examines the differences and potential links with an interesting recent proposal based on a parameterized family of means, also for extending analogical proportions to numerical values. 
\end{itemize}
\section{Piaget's logical proportion and analogy}
\label{Piaget}
Let $a$, $b$, $c$, $d$ denote four logical propositions.  Then  $a$, $b$, $c$, $d$ make a  {\it logical proportion} in the sense of Piaget \cite{Piaget1952,Piaget1953} if the two following conditions hold $$a \wedge  d = b \wedge c \mbox{ and  }a \vee d = b \vee c.$$
An analogical proportion is a statement involving four items, usually denoted $a\!:\! b\!::\! c\!:\!d$, which reads ``$a$ is to $b$ as $c$ is to $d$''.  When $a$, $b$, $c$, $d$  are propositional truth values, the two conditions underlying Piaget's logical proportions define the following quaternary connective    

$$Pia(a, b, c, d) =((a \wedge d) \equiv (b \wedge c)) \wedge ((  a \vee   d) \equiv (  b \vee c)) =((a \wedge d) \equiv (b \wedge c)) \wedge ((\neg a \wedge \neg d) \equiv (\neg b \wedge \neg c))$$

This connective when it is true emphasizes \emph{similarity} and can be read as  ``what $a$ and $d$ have in common (positively or negatively), $b$ and $c$ have it also''. It also says that ``when $a$ and $d$ differs (one is true, the other is false) then  $b$ and $c$ also differ''.  

$Pia(a, b, c, d)$  is true only for the 6 patterns exhibited in the table below (and thus false for the $2^4 - 6 = 10$ remaining patterns).
$$
\begin{array}{cccccc}
a & b & c & d  \\
\hline
\hline
 0 & 0 & 0 & 0  \\
\hline
1 & 1 & 1 & 1  \\
\hline
0 & 0 & 1 & 1  \\
\hline
1 & 1 & 0 & 0  \\
\hline
0 & 1 & 0 & 1 \\
\hline1 & 0 & 1 & 0 \\
\hline
\end{array}
$$

 The above logical expression is known to be one of the expressions of a Boolean analogical proportion \cite{MicPraECSQARU2009}. Another equivalent expression emphasizes \emph{dissimilarity}  inside pairs $(a, b)$ and $(c, d)$. Namely we have :

$$a:b::c:d = ((a \wedge \neg b)  \equiv (c \wedge \neg d)) \wedge ((\neg a \wedge b)  \equiv (\neg c \wedge d))) \ \  \ \ (1)$$

 It precisely expresses that ``$a$ {differs} from {$b$} as  {$c$} {differs} from   {$d$} and {$b$} {differs} from {$a$} as {$d$} {differs} from {$c$}'' (and ``when {$a$} and {$b$} do not differ, {$c$} and {$d$} do not differ'').  Thus we also have 
$$a:b::c:d =Pia(a, b, c, d)=((a \wedge d) \equiv (b \wedge c)) \wedge ((  a \vee   d) \equiv (  b \vee c)) \ \ \ \ (2)$$
It can be easily checked that $a:b::c:d$ satisfies the postulates usually assumed  for analogical expressions:  
-\begin{itemize}
   \item \emph{reflexivity}: $a:b:a:b$;
    \item \emph{symmetry}: $a:b::c:d \Rightarrow c:d::a:b$ ;
    \item {stability under \emph{central permutation}}: $a:b::c:d \Rightarrow a:c::b:d$. 

 \end{itemize}
 
As a consequence, an analogical proportion
also satisfies \begin{itemize}
    \item $a:a::b:b$ (sameness) ;
    \item   $a:b::c:d$  $\Rightarrow d:b::c:a$ ({external permutation}); 
  \item $a:b::c:d \Rightarrow b : a :: d : c$ (internal reversal);
  \item $a:b::c:d \Rightarrow d : c :: b : a$ (complete reversal).

\end{itemize}

Note that reflexivity postulate $a:b:a:b$ , true for all $a$ a d $b$, forces a Boolean analogical proportion to be true for the valuations $(0, 0, 0, 0)$, $(0, 1, 0 , 1)$, $(1, 0, 1, 0)$, and $(1, 1, 1, 1)$. The minimal Boolean model that satisfies the three above postulates is true for the 6  patterns for  Boolean variables $a, b, c, d$ exhibited in the above table 
\cite{PraRicIJAR2018}. 

Besides it can be checked that Boolean analogical proportions are transitive:
$$(a:b::c:d) \wedge (c:d::e:f) \Rightarrow a:b::e:f.$$

Analogical proportions have been extended to numerical values $a, b, c, d$ (assumed to be normalized between 0 and 1) by starting from (1) or (2)  and by replacing connectives $\wedge$, $\vee$, $\neg$, $\equiv$ by multiple-valued connectives, the choice of these latter connectives being dictated by the preservation  of a maximum of desirable properties \cite{PraRicISMVL2010,DubPraRicFSS2016}. This has led in (1)  to take i) $\min$ for the central $\wedge$, ii) $\min(s \to_ {L} t,t\to_{L} s)=1-|s-t|$
 for the two equivalence connectives, where 
$s \to_{L} t = \min(1, 1 - s + t)$ is \L ukasiewicz implication,  iii) the bounded difference $\max(0, s - t) = 1 -  (s\to_{L} t)$,  (using $1 - (\cdot)$ as negation) for the four expressions of the form $s \wedge \neg t$.  For (2), one has kept the same choices for the central $\wedge$ and for the equivalences, and used $\min$ and $\max$ for the inside $\wedge$ and $\vee$.  

This graded view leads to privilege the connectives of MV algebras. In what follows, we explore another avenue.
 
\section{A triangular norm-based definition of an analogical proportion}
\label{triang}
Triangular norms are associative operations of the unit interval \cite{SS63}, which have been extensively studied in fuzzy set theory \cite{KMP00}, triangular norms being the proper operations for defining fuzzy set intersections, and their dual co-norms for  defining fuzzy set unions. We start with a brief refresher on triangular norms, before introducing them in a new definition of analogical proportions for  numerical values.

\subsection{A brief refresher on triangular norms}
 Triangular norms (t-norms for short) and triangular conorms (t-conorms for short) were invented by Schweizer and Sklar \cite{SS63,KMP00}, in the framework of probabilistic metric spaces, for the purpose of expressing the triangular inequality. They also turn out to the most general binary operations on [0, 1] that meet natural and intuitive requirements for conjunction and disjunction operations. Namely, a    t-norm $T$ is a binary operation on $[0, 1]$, i.e., $T: [0, 1] \times: [0, 1] \to [0, 1]$, that satisfies the following conditions: 
\begin{itemize}
\item commutative: $T(a, b) = T(b, a)$;
\item  associative: $ T(a, T(b, c)) = T(T(a, b), c)$;
\item non-decreasing in both arguments: $T(a, b)  \leq T (a', b')$ if $a  \leq a'$ and $b  \leq b'$;
\item boundary conditions:  $T (a, 1) = T (1, a) = a.$ 
\end{itemize}
It can be proved that $T(a, 0) = T(0, a) = 0$.   It is known that the minimum operation is the greatest t-norm, i.e., for any t-norm $T$, $T(a, b)  \leq \min(a, b)$ holds for all $a, b \in [0, 1]$. Typical basic examples of t-norms are 
\begin{itemize}
\item the minimum : $T(a, b) = \min(a, b)$, 
\item the product: $ T(a, b) = a \cdot b$,
\item  the linear (or {\L}ukasiewicz) t-norm: $T(a, b) = \max(0, a + b - 1)$.
\end{itemize} 
Note the inequalities:  $\max(0, a + b - 1)  \leq a \cdot b  \leq \min(a, b).$

These three cases are important since any {\em continuous} {t-norm} is definable as  an {\em ordinal sum\/}\footnote{An ordinal sum is defined in the following way.  Let $ \{ \mathcal{I}_n \}_n$ be  a countable family of sub-intervals in [0, 1]. To each  $\mathcal{I}_n=[x_n, y_n] $,  we associate  an Archimedian t-norm $T_n$. Then  the ordinal sum is the triangular  norm $T(a, b) = x_n +  (y_n - x_n)T_n (\frac{a- x_n}{y_n - x_n},  \frac{b - x_n}{y_n - x_n})$ if $a, b \in [x_n, y_n]$; $T(a, b) = \min(a,b)$ otherwise \cite{KMP00}.}  of copies of \L ukasiewicz, 
minimum and product t-norms (see, e.g,. \cite{KMP00}).

In the following, we only consider continuous t-norms.  A t-norm is said to be Archimedian if $\forall a, 0< a <1, T(a,a) < a $.  Any continuous Archimedian t-norm has an additive generator $f$, and can be written as
$T(a, b) = f^{(-1)} (f(a)+f(b))$
where $f$ is a strictly decreasing function from [0, 1] to $[0, +  \infty)$  and $f^{(-1)}$ is its pseudo-inverse defined by 
$f^{(-1)}(x) =1$ if $x \in [0, f(1)]$,  $f^{(-1)}= f^{-1}(x)$ if  $x \in [f(1), f(0)]$ , $f^{(-1)}= 0$ if $ x \in[f(0), +  \infty)$.  If $f$ is such that $f(1) =0$ and $lim _ {x \to 0^+}f(x) = + \infty$ , the t-norm is said to be {\it strict}. The product is a typical example of strict t-norm (for which $f(x)= - ln(x)$). When $T$ is strict, $f^{(-1)}(x) =f^{-1}(x) $.  If  $f(1) =0$ and $f(0)$ is finite, the t-norm is said to be {\it nilpotent};  the {\L}ukasiewicz) t-norm is a typical example ($f(x) = 1 -x$).

The De Morgan-like dual notion of a t-norm (w.r.t. negation $n(a) = 1 - a$, or a more general strong negation) is that of a    t-conorm $S(a,b)= n(T(n(a), n(b)))$ (a strong negation is such that $n(0)  = 1$; 
	$n(1)  = 0$; $n(a)  \geq n(b),  \mbox{~if~} a \leq b$,  and $n(n(a)) = a$; 
in the following  we use $n(a) = 1 - a$, for all $a \in [0, 1]$). 

Thus, a binary operation $S$ on [0, 1] is called a t-conorm if it satisfies the same properties as the ones of a t-norm except for the boundary conditions, namely, here $0$ is an identity and $1$ is absorbent:
	\begin{center} 
boundary conditions:  $S (0, a) = S (a, 0) = a. $
\end{center}
	
	Hence $S (a, 1) = S(1, a) = 1$. Dually, the maximum operation is the smallest t-conorm ($S(a, b) \geq \max(a, b)$). 
Typical basic examples of t-conorms are the dual of minimum, product and {\L}ukasiewicz' t-norms, namely the maximum $S(a, b) = \max(a, b)$, the so-called probabilistic sum $S(a, b) = a + b - a · b$  and the bounded sum $S(a, b) = \min(1, a + b)$. Note now the inequalities
	$\max(a, b)  \leq a + b - a \cdot b  \leq \min(1, a + b).$

\subsection{Definition  and properties}
\begin{definition}
Let $a, b, c, d $ be four numerical values in [0, 1].  $a, b, c, d $ are said to satisfy an analogical proportion based on the t-norm $T$ (and its dual t-conorm $S_T$), which is denoted 
$A_T(a, b, c, d)$, if and only if $ T(a, d) =T(b,  c)$ and $S_T(a, d) = S_T( b, c))$.
\end{definition}
Note that this definition is all-or-nothing: $A_T(a, b, c, d)$ is true or false.  This contrasts with the multiple-valued logic extensions of $a : b :: c : d$, where the multiple-valued quaternary connective obtained becomes a matter of degree when $a, b, c, d $ are numerical values \cite{DubPraRicFSS2016}.  Definition 1 might be weakened by requiring only equality for $T$ (or for  $S_T$),  but in the following we keep it as it is for symmetry reason. 

It is easy to check that

\begin{itemize}
\item when $a, b, c, d \in \{0, 1\} $, $A_T(a, b, c, d)$ is true only for the 6 valuations of the table in Section  \ref{Piaget} that make $a : b :: c : d$ true.
\item $A_T(a, b, c, d)$ satisfies the three postulates of  analogical proportions:
i) reflexivity, ii) symmetry, and iii) stability under central permutation, thanks to the commutativity of  $T$  (and thus of $S_T$). It is clear that  $A_T(a, b, c, d)$ also satisfies the consequences of the three postulates.
\end{itemize}

\begin{example}Let us consider the examples of the three main t--norms.
\begin{itemize}
    \item $T(a, b) = \min(a,b) $ and $S_T(a, b) = \max(a,b)$.  Then  $A_T(a, b, c, d)$ means $\min(a,d)$ $= \min (b, c)$ and $\max(a, d)= \max(b, c)$.  So $A_T(a, b, c, d)$ holds only if $a=b$ (resp. $a=c$) and then $c=d$ (resp. $(b = d)$.
    \item  $T(a, b) = ab$ and $S_T(a, b) = a + b - ab$.  Then  $A_T(a, b, c, d)$ means    $ad =bc$ and  $a + d - ad = b + c - bc$, which is equivalent to $ad =bc$ and $a+d = b + c$. 
    Thus, $a, b, c,  d$ form both an arithmetic proportion (i.e.,  $a-b = c- d$) , and a geometric proportion  (i.e., $\frac{a}{b}=\frac{c}{d} $ provided that $b \neq 0$ and $d \neq 0$).
In the case $a=0$, this imposes $b$ or $c$ to be 0 as well. In case $b=0$, then $d=c$, if $c=0$, then $d=b$. This leads to the only candidate proportions $A_T(0, 0, c, c)$ and
 $A_T(0, b, 0, b)$. More generally,  letting  $a+d=S$ and $ad=P$, we   obtain the equation  $a^2 - Sa  + P$ which has two solutions $a=b$ (and then $c=d$) and $a=c$ (and then $b=d$). 
    \item  $T(a, b) = \max(0, a+b -1)$ and $S_T(a, b) = \min(1, a +b)$.  Then  $A_T(a, b, c, d)$ means $\max(0, a+d -1)= \max(0, b +c -1)$ and  $\min(1, a +d)= \min(1, b + c)$, i. e. ,  $a+ d = b +c$ , or if we prefer  $a - b = c - d$.
    \end{itemize}
    \end{example}
  
Thus, we obtain two different views of an analogical proportion between numerical values:     \begin{itemize} 
\item a {\it conservative} view with $T$ equal to minimum or product, where  the only accepted analogical proportions are of the form $a: a :: c : c$ or  $a: b ::a :b$.
\item a {\it liberal} view with $T$ equal to {\L}ukasiewicz) t-norm where  $a : b ::c : d$ holds if  and only if $a-b = c- d$. 
\end{itemize}
    
These two options  corresponds to the cases where the two multiple-valued logic extensions of $a : b :: c : d$ \cite{DubPraRicFSS2016} are equal to 1 respectively.  It is clear that the conservative view is much more drastic than the liberal one.

Besides, it is clear that $A_T(a, b, c, d)$  is transitive (i.e.,$A_T(a, b, c, d),  A_T(c, d, g, h) \Rightarrow A_T(a, b, g, h) $) 
when $T$  is one of the three main triangular norms. But the property is more general. For instance, when $T$ is strict,  using its additive generator, we have $f(a) + f(d) = f(b) + f(c)$ and $f(c) + f(h)= f(d) + f(g)$ that entails $f(a) + f(h)= f(b) + f(g)$, and similarly for the  dual co-norm equality.


\subsection{Appropriateness of Frank's triangular norms}
There exist several parameterized families of triangular norms \cite{KMP00}.  One is the  Frank's family  \cite{Frank79}, which includes the main triangular norms. There are defined by

$$
T_F^p(a,b) = \left\{
    \begin{array}{ll}
       \min(a, b) & \mbox{if } p=0 \\
        a \cdot b & \mbox{if } p=1 \\
       \max(0, a + b-1) & \mbox{if }p=  +\infty  \\
        \log_p(1 + \frac{(p^a - 1)(p^b -1)}{p - 1}) & \mbox{otherwise.}
    \end{array}
\right.
$$
and their additive generator is 
\vspace{-0.4cm}
$$
f_F^p(x) = \left\{
    \begin{array}{ll}
        - \log x & \mbox{if } p=1 \\
       1 -x & \mbox{if }p=  +\infty  \\
        \log(\frac{p - 1}{p^x - 1 }) & \mbox{otherwise.}
    \end{array}
\right.
$$
More importantly for our topic,  Frank's triangular norms  (and their ordinal sums) are  the only triangular norms  that satisfy the following 
remarkable property
$$\forall p,  T_F^p(a,b) + S^p_{T_F}(a,b) =  a + b  \ \ \ \ (3)$$
This echoes the following canonical analogical proportion \cite{PraRicLU2013} that holds in any lattice structure \cite{BarbotMPAIJ19}:
\vspace{-0.2cm}$$a \wedge b : a :: b : a \vee b $$
since it is clear that (3) expresses that $T_F^p(a,b) : a :: b : S^p_{T_F}(a,b) $ is an arithmetic proportion.

{ 
A candidate option to estimate a $p$ such that $A_{T^p_{F}}(a,  b, c, d)$ is to minimize the sum of the difference
$|T^p_{F}(a, d)-T^p_{F}(b,  c)+ |S^p_{F}(a, d) - S^p_{F}( b, c))|$. As a preliminary,
we have experimented with $p=2$, $p=10$ and $p=100$, with fixed values for $a,b,c$ with $a \leq b \leq c \leq d$: e.g.,  $a = 0.01, b = 0.2, c = 0.3$. 
In Fig. \ref{figure1}, we provide the curves of  
$diff_p(x)=|T^p_{F}(a, x)-T^p_{F}(b,  c)+ |S^p_{F}(a, x) - S^p_{F}( b, c))|$, for $p=2, 10, 100$ and $x$ varying linearly from $0.3$ to $1$. We observe that the value $x=d$ such that $diff_p(x)=0$ slowly increases with $p$ to  0.49 (obtained for $p= + \infty$, i.e., for $a + d = b +c$).
\begin{figure} 
\vspace{-0.5cm}
    \centering
    \subfigure{\includegraphics[width=0.30\textwidth]{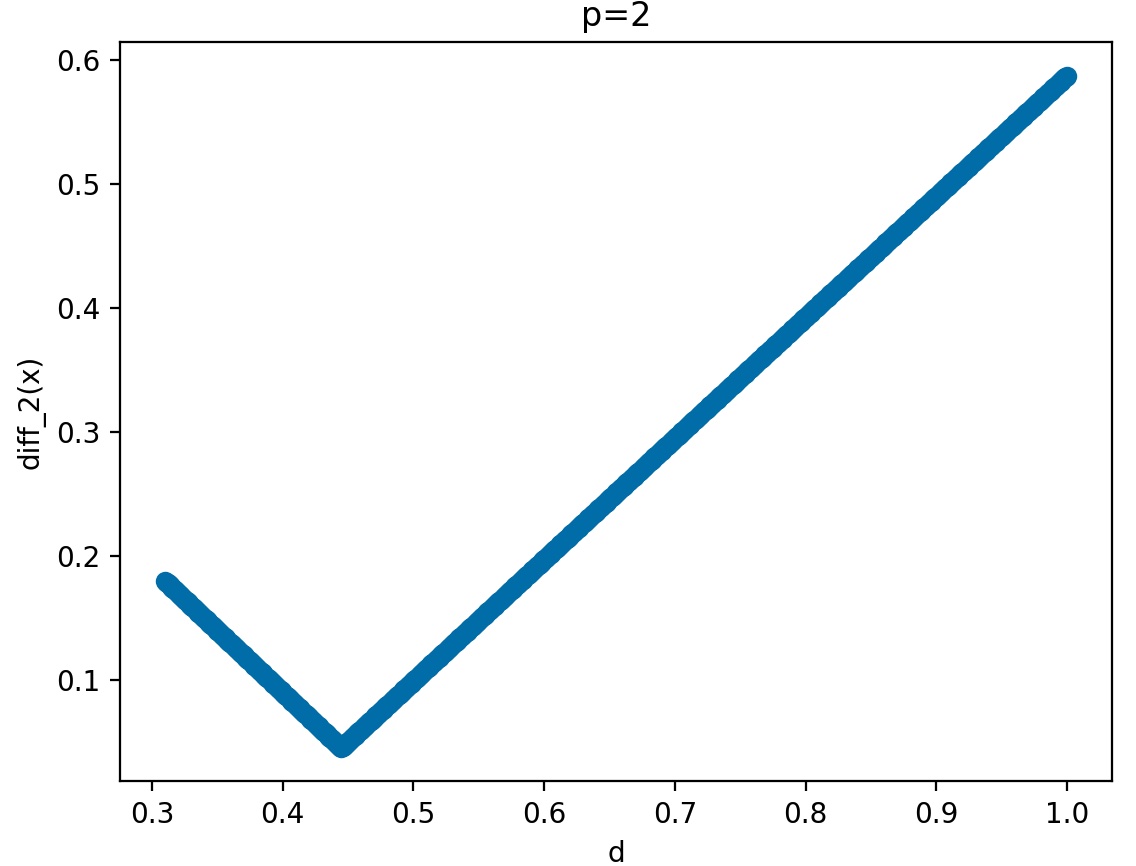}}
    \subfigure{\includegraphics[width=0.30\textwidth]{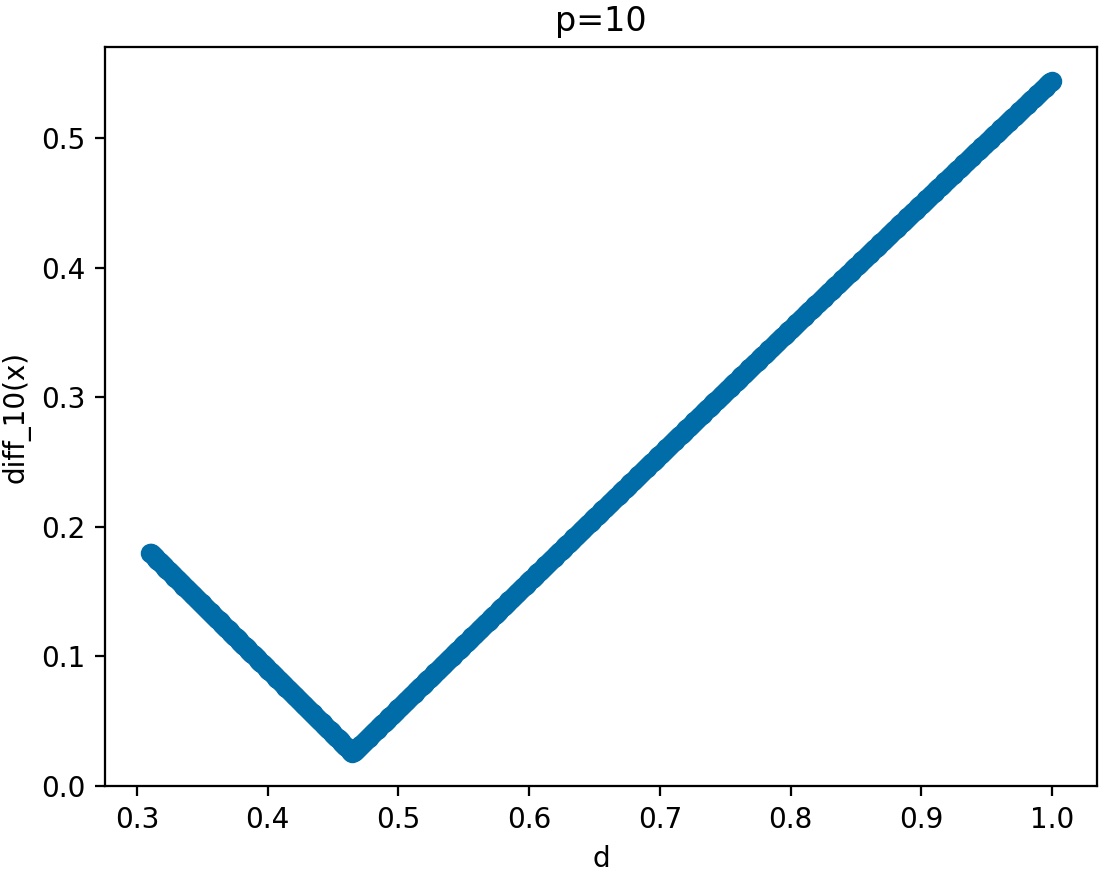}}
    \subfigure{\includegraphics[width=0.30\textwidth]{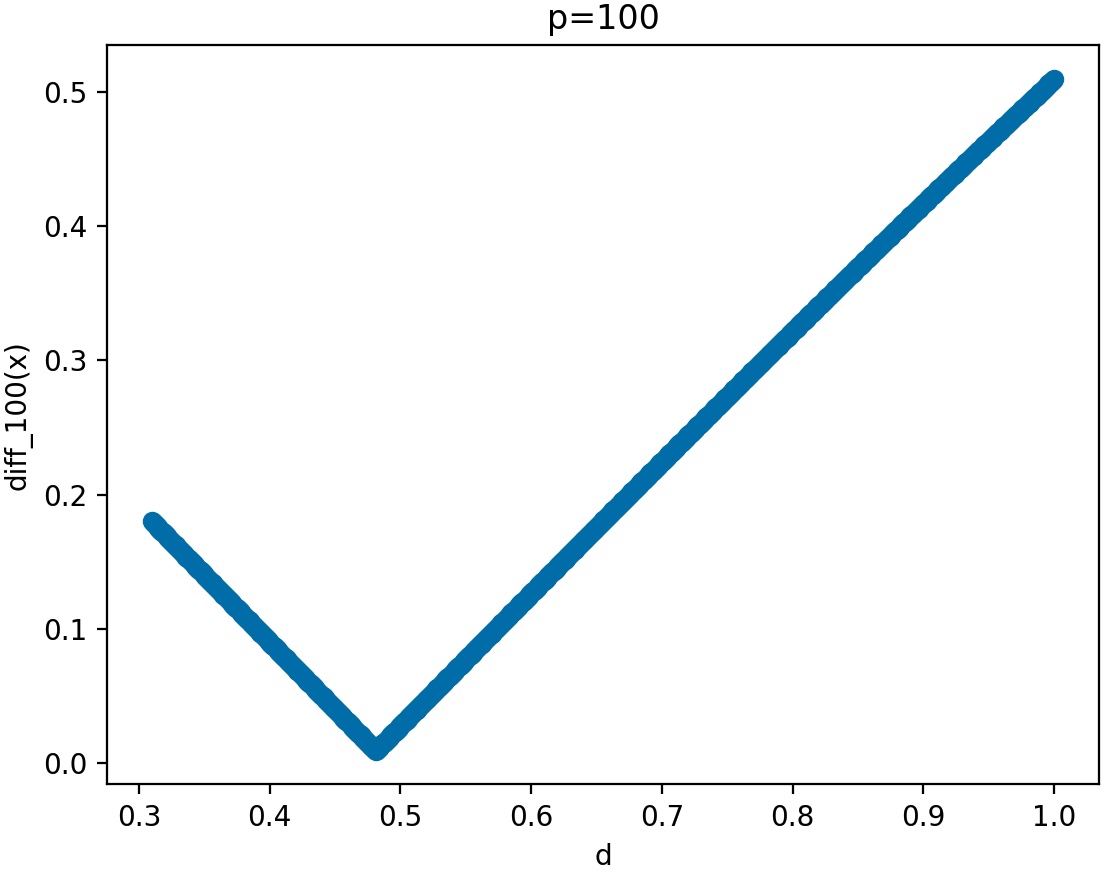}}
    \vspace{-0.4cm}\caption{$p=2$ -  $p=10$ -  $p=100$ }
    \label{figure1} 
\end{figure}
}

\section{Related work and discussion} \label{related} 
Very recently, in \cite{LepCou2024}, a nice proposal has been made for defining  an analogical proportion between numerical values, using generalized means.  We assume, without loss of generality, that we still deal with numerical values $a, b, c, d$ in the unit interval $[0, 1]$. 
\begin{definition}
Let $a, b, c, d  \in[0, 1]$.  $a, b, c, d $ are said to satisfy an analogical proportion if and only if $M_r(a,d) = M_r(b,c) $, 

\noindent where $M_r(x, y) = \frac{1}{2} (x^r  + y^r)^{1/r}$. 
\end{definition}For  $r= -1, 0, 1, 2$,  we obtain the harmonic, geometric, arithmetic, and quadratic mean respectively.  We have $\forall r, \min(x,y) \leq  M_r(x, y) \leq \max(x, y)$, and the bounds are reached for  $r=-\infty $ and $r=+\infty$  respectively. Moreover,  it is established \cite{LepCou2024} that  given $a \leq b \leq c \leq d $,  $\exists !r  \mbox{  such that  }  M_r(a,d) = M_r(b,c) $.

There seems to be an obvious parallel between Definition 1 and Definition 2, up to the point that the conditions are respectively stated in terms of  t-norms and dual t-conorms on the one hand and in terms of mean on the other hand.  The  two definitions coincides  if  $T$ is the {\L}ukasiewicz t-norm and  $M_p = M_1$. Definition 1 is closer to logic and to multiple-valued logic extensions of analogical proportion  \cite{DubPraRicFSS2016} than Definition 2 that uses means which are not multiple-valued logic operations strictly speaking. 

The use of the parameterized family of  Frank's t-norms enables us to modulate the conditions defining $A_{T^p_{F}}(a,  b, c, d)$;  we may wonder  if  when $0 \leq a \leq b \leq c \leq d \leq 1,  \  \exists p \mbox{ such that } A_{T^p_{F}}(a,  b, c, d) \mbox{ holds}.$
This is an open question. At least we know  that the solution may not be unique, since, e.g., $A_{T^p_{F}}(0.5, 0.5,, 0.7, 0.7) $ holds  for  $p= 0, 1, + \infty$, (i.e., for $T= \min$, product, and {\L}ukasiewicz) t-norm. 

Besides,  a long time ago, a multiple-valued logic extension of (1) based on 
Goguen's implication  ($a \to b = 1$ if $a =0$  and $a \to b = \min(1, b/a)$ otherwise) was proposed  \cite{MicPraECSQARU2009} which is equal to 1 if and only if $ad= bc$, thus coinciding with  Definition 2 for  $M_0$. This suggests that the agreement of Definition 2 with logic may be greater than it appears at first glance.

\section{Concluding remarks}
The ambition of this note is to show the possibility of a definition, based on t-norms,
of an analogical proportion between numerical values that makes sense and merits to be studied further.  Some families of t-(co)norms are closer to generalized averages than Frank's, such as Schweizer and Sklar's family \cite{SS63,KMP00},  also including the main t-norms, but they do not satisfy (3). Further research should lead to a unifying view on the topic. 

\section*{Acknowledgements} This research was supported by the ANR project ``Analogies: from Theory to Tools and Applications'' (AT2TA), ANR-22-CE23-0023.


\end{document}